\documentclass[11pt]{article}

\usepackage[utf8]{inputenc}
\usepackage[T1]{fontenc}
\usepackage[english]{babel}
\usepackage{amsmath,amssymb}
\usepackage{graphicx}
\usepackage{booktabs}
\usepackage{array}
\usepackage{float}
\usepackage{geometry}
\usepackage{tikz}
\usetikzlibrary{arrows.meta,positioning,shapes.geometric}
\usepackage[colorlinks=true,linkcolor=blue,citecolor=blue,urlcolor=blue]{hyperref}
\usepackage{natbib}
\usepackage{xcolor}
\usepackage{caption}

\renewenvironment{abstract}{\par\noindent{\bfseries\abstractname}\par\smallskip\noindent}{\par\medskip}

\newcommand{\compass}{COMPASS}
\newcommand{\agent}{\texttt{ComplianceAgent}}

\begin{document}

\begin{flushleft}
{\LARGE\bfseries An LLM-Based Agent for Legal Compliance Assessment: Architecture, Memory, and Evaluation in a Canadian and Quebec Context\par}
\vspace{1.2em}
{\large Soukaina Alaoui Ismaili and Jean-S\'ebastien Dessureault\par}
\vspace{0.4em}
Department of Electrical and Computer Engineering \\
Universit\'e du Qu\'ebec \`a Trois-Rivi\`eres (UQTR), Trois-Rivi\`eres, Qu\'ebec, Canada\par
\vspace{0.3em}
\texttt{Soukaina.Alaoui.Ismaili@uqtr.ca}, \texttt{Jean-Sebastien.Dessureault@uqtr.ca}\par
\vspace{0.3em}
October 2026
\end{flushleft}
\vspace{0.5em}

\begin{abstract}
Checking whether a text complies with a body of law is slow when done by hand and fragile when done with keyword rules. This paper presents \agent{}, the legal compliance sub-agent of \compass{}, a multi-agent framework that evaluates texts along four axes: digital sovereignty, carbon impact, compliance, and ethics. The agent relies on a large language model that scores a text on a fixed rubric, grounds its judgment in a reference legal corpus, and, when the score falls at or below a critical threshold, runs a corrective chain made of self-healing rewrites, a compliance label, an explainability report, and escalation to a human reviewer. We describe what the language model does at each step, how the memory of the agent is organized, and how the same implementation can be pointed at different reference corpora. We evaluate the agent in a Canadian and Quebec context, on a corpus of thirteen legal instruments that brings together federal and provincial texts, international treaties, and municipal documents of the City of Trois-Rivi\`eres. The evaluation relies on a hand-annotated gold standard of fifteen texts and combines classification metrics, a repeated-query stability test, retrieval precision, and a composite score. On this gold standard the agent separates every conforming text from every non-conforming one, its scores remain stable under repeated queries, and the index returns relevant passages.
\end{abstract}


\newpage
\section{Introduction}
\label{sec:intro}

Organizations that build or deploy automated systems are increasingly asked to show that their practices respect the rules that apply to them. In Canada and Quebec these rules come from several levels at once: federal statutes such as the Personal Information Protection and Electronic Documents Act \citep{canada2000pipeda}, provincial statutes such as Quebec's Law~25 \citep{quebec2021loi25}, municipal policies and by-laws, and the international treaties that frame them. Voluntary frameworks such as the NIST AI Risk Management Framework \citep{nist2023airmf} and management system standards such as ISO/IEC 42001 \citep{iso42001} add further expectations. All of this turns into documents, statements, and policies that someone has to read and judge. Doing so by hand does not scale with the volume of text, and rule-based checkers break as soon as a statement does not match a fixed pattern. A language model that can read a statement in context and compare it with the relevant provisions offers a practical middle path, provided that its judgment can be tied to the source text and explained to a human reader.

Using a language model in this role raises questions that are rarely answered together. What exactly does the model decide, and what is left to the surrounding code? What does the system remember, and what does it retrieve at the moment of judgment? How should its output be evaluated when no accepted benchmark exists for the legal context of interest? Work on legal language processing tends to evaluate models on reasoning tasks, and work on language agents tends to describe memory and refinement loops in general terms. An agent that scores, corrects, explains, and escalates over a given legal corpus is seldom documented in enough detail to be reproduced and criticized. We organize the study around three research questions. The first concerns design: how can the roles of the model, of retrieval, and of memory be organized in a compliance agent so that each can be described and tested? The second concerns performance: how well does the agent separate conforming from non-conforming texts in a Canadian and Quebec legal corpus, and how stable are its scores? The third concerns evidence: what can a small evaluation made by a single annotator actually support?

This paper describes \agent{}, the compliance sub-agent (COM) of \compass{}, a research framework \citep{dessureault2026compass} that evaluates a text along four independent axes: digital sovereignty (SOV), carbon impact (CAR), regulatory compliance (COM), and ethics (ETH). Each axis is handled by a dedicated sub-agent built on a shared base class, and an orchestrator sends a text to the relevant sub-agents, either on explicit request or through a filter that is itself a language model call. The work reported here concerns the compliance axis only. The other sub-agents are developed by other members of the project and fall outside the scope of this paper.

The agent is configurable by domain, which means that the reference corpus is a parameter and not a path written into the code. In this study it is pointed at a corpus of thirteen Canadian, Quebec, municipal, and international legal instruments. This paper makes three contributions. First, we document the architecture precisely enough for a reader to know what the language model does, what each kind of memory holds, and what the retrieval step does and does not contribute. Second, we define the evaluation metrics mathematically and apply them to a hand-annotated gold standard. Third, we report what the evaluation shows and, just as importantly, what it cannot show given its size and design.

The rest of the paper follows the usual structure of a scientific article. Section~\ref{sec:related} reviews the state of the art. Section~\ref{sec:methodology} presents the methodology, namely the architecture of the agent, the language model and its prompts, the data, and the evaluation protocol and metrics. Section~\ref{sec:results} reports the results. Section~\ref{sec:discussion} discusses them, with their limits and the perspectives they open, and Section~\ref{sec:conclusion} concludes.

\section{State of the Art}
\label{sec:related}

\paragraph{Retrieval-augmented generation.} Retrieval-augmented generation \citep{lewis2020retrieval} addresses the fact that the knowledge stored in a model's parameters is static and sometimes wrong, by conditioning generation on passages fetched from an external corpus. Surveys such as \citet{gao2023ragsurvey} describe the many ways in which indexing, retrieval, and prompting can be combined. In legal settings the cost of ungrounded output is high: \citet{dahl2024legalfictions} show that general-purpose models hallucinate frequently on legal queries, which is one reason to tie a compliance judgment to a retrievable source. Our retrieval layer uses MPNet-based sentence embeddings \citep{song2020mpnet,reimers2019sentencebert}, a persistent open-source vector store \citep{chromadb}, and a composable application framework \citep{langchain}.

\paragraph{Language models as evaluators.} Using a language model to grade text against a rubric has been studied under the name LLM-as-a-judge \citep{zheng2023judging}. That line of work mostly concerns open-ended generation quality and pairwise preference. \agent{} applies the same principle to a narrower task: it scores one text against a defined body of law with a fixed rubric, and the result is compared with a human annotation. This framing makes ordinary classification metrics applicable, in contrast with open-ended judgment where no gold label exists.

\paragraph{Legal language processing.} Domain-specific pretrained models such as LEGAL-BERT \citep{chalkidis2020legalbert} and benchmarks such as LegalBench \citep{guha2023legalbench} evaluate legal reasoning on tasks like classification, rule application, and issue spotting. They do not target the end-to-end behavior of an agent that retrieves, scores, rewrites, and escalates, which is the object of study here.

\paragraph{Agent memory and iterative refinement.} The cognitive architecture for language agents proposed by \citet{sumers2024cognitive} distinguishes working memory from long-term stores, themselves divided into episodic, semantic, and procedural memory. We use this vocabulary in Section~\ref{sec:memory} to say plainly what our agent has and does not have. Our self-healing loop belongs to the family of iterative refinement methods, such as Self-Refine \citep{madaan2023selfrefine} and Reflexion \citep{shinn2023reflexion}, in which a model's output is critiqued and revised. In our case the critique is the explanation produced by the scoring step and the stopping test is the compliance threshold. The compliance label we produce plays a role comparable to a model card \citep{mitchell2019modelcards}: a short structured document that states what was evaluated, with what outcome, and on what sources.

\paragraph{Positioning.} Taken together, these lines of work provide the components of our agent: retrieval over a corpus, a model that grades against a rubric, a memory vocabulary, and iterative refinement. What they leave open is how these components are assembled into one compliance agent, what each of them contributes at the moment of judgment, and how such an agent behaves on a Canadian and Quebec legal corpus.

\section{Methodology}
\label{sec:methodology}

This section describes how the agent is built and how it is evaluated. We first present its architecture, then the language model and its prompts, the data used for the evaluation, and finally the evaluation protocol and the metrics.

\subsection{System Architecture}
\label{sec:architecture}

\subsubsection{The \compass{} Framework}

\compass{} was introduced by \citet{dessureault2026compass}. It is built from four sub-agents that inherit from a common \texttt{SubAgent} base class. The base class provides what all sub-agents share: reading prompt templates from disk, calling the language model, extracting a JSON object from the model's reply, and building a vector index from a PDF. An orchestrator, \texttt{COMPASS\_Agent}, instantiates the four sub-agents and sends an incoming text to those that are relevant. The compliance sub-agent is the object of this paper.

\begin{figure}[H]
\centering
\begin{tikzpicture}[box/.style={draw, rounded corners, align=center, minimum height=0.9cm, inner sep=4pt, font=\small},
  arr/.style={-{Latex[length=2mm]}, thick}]
\node[box] (in) at (0,0) {Input\\text};
\node[box] (filter) at (3.4,0) {Filter\\(language model)};
\node[box] (sov) at (7.0,1.65) {SOV};
\node[box] (car) at (7.0,0.55) {CAR};
\node[box, very thick] (com) at (7.0,-0.55) {COM};
\node[box] (eth) at (7.0,-1.65) {ETH};
\node[box] (out) at (10.6,0) {Scores and\\explanations};
\draw[arr] (in) -- (filter);
\foreach \n in {sov,car,com,eth}{\draw[arr] (filter.east) -- (\n.west); \draw[arr] (\n.east) -- (out.west);}
\draw[arr, dashed] (in.north) -- ++(0,2.3) -| (sov.north);
\node[font=\scriptsize, above] at (3.4,2.75) {explicit choice of a sub-agent};
\node[font=\scriptsize, align=center] at (3.4,-1.0) {zero, one, or several\\sub-agents selected};
\end{tikzpicture}
\caption{Architecture of \compass{} with the filter. The filter asks a language model which sub-agents are relevant to the input text, so that only those are run. An explicit choice of a sub-agent bypasses it (dashed arrow). The compliance sub-agent (COM, bold outline) is the subject of this paper.}
\label{fig:compass}
\end{figure}
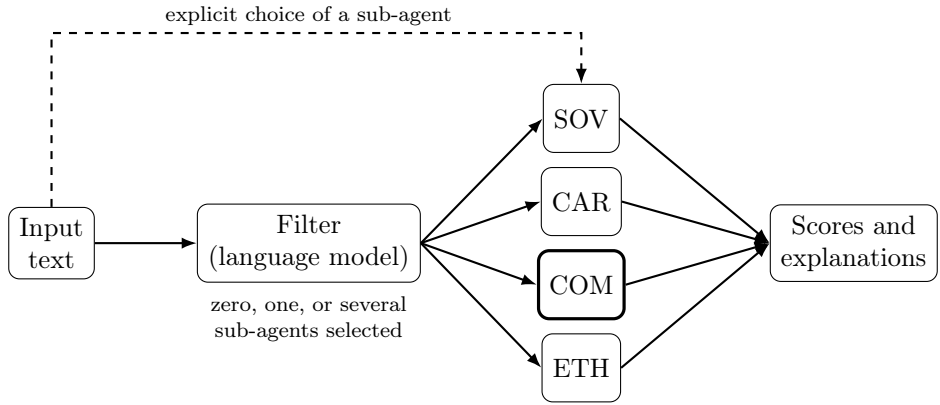

The version of the framework used here adds a feature that was not part of the initial paper: a filter. Before any sub-agent runs, the orchestrator questions a local language model about the subject of the input text, with a dedicated prompt that lists the four modules and asks for a bare list of module keys. The model decides which sub-agents to call: one, several, or none. The reply is parsed as a list, keys that do not name an existing sub-agent are discarded, and a reply that cannot be parsed selects no sub-agent. Only the selected sub-agents then analyze the text, which avoids calling all of them on every input and keeps the calls to the relevant ones. A caller can bypass the filter by naming a sub-agent explicitly. Figure~\ref{fig:compass} shows the resulting architecture. The filter is described here for completeness, but it is not evaluated in this paper, whose experiments address the compliance sub-agent alone.

\subsubsection{The \agent{} Sub-Agent}

\agent{} exposes two operations. The method \texttt{analyze}$(x, \text{rag})$ takes a text $x$ and returns a score $s \in [0,1]$ together with a short explanation. When the model's reply cannot be parsed, or when the model answers that the information is insufficient, no score is returned and the code stores the value \texttt{None}. The method \texttt{act}$(x)$ rescores the text and, if $s \le \tau$ with the critical threshold $\tau = 0.80$, triggers the corrective chain of Section~\ref{sec:action-chain}. This two-level design deliberately mirrors the ethics sub-agent of the same framework, so that the user meets the same interaction pattern on every axis.

\subsubsection{The Role of the Language Model}
\label{sec:llm-role}

No step of the agent is implemented by hand-written scoring logic. A single general-purpose language model is called with a different instruction at each step, and the instructions live in prompt files (Table~\ref{tab:llm-roles}). The model acts as a scorer in \texttt{analyze}, as an editor in \texttt{self\_healing}, as a technical writer in \texttt{compliance\_label} and \texttt{explainability}, and as a case officer in \texttt{human\_in\_the\_loop}. One further call, executed once when the agent is constructed, makes the model a summarizer of the reference corpus (Section~\ref{sec:memory}).

Only the scoring step has a strict output contract. Its prompt demands a JSON object with the keys \texttt{score} and \texttt{explanation}, and the base class extracts the substring between the first opening brace and the last closing brace and parses it. If parsing fails, or if the model answers N/A, no score is returned and the explanation carries an error message. The other steps return free text. Because the score is the only quantity on which the control flow depends, this single contract is enough to keep the pipeline composable.

\begin{table}[H]
\centering
\small
\begin{tabular}{@{}p{2.9cm}p{2.1cm}p{4.6cm}p{4.4cm}@{}}
\toprule
Step & Role & Input to the model & Output \\
\midrule
Memory build (once) & Summarizer & Top 3 passages for a fixed query & Corpus summary $m$ \\
\texttt{analyze} & Scorer & Text $x$, summary $m$, rubric & JSON with score and explanation \\
\texttt{self\_healing} & Editor & Current text, explanation of the failure & Rewritten text \\
\texttt{compliance\_label} & Technical writer & Score, explanation, 3 retrieved excerpts & Structured label \\
\texttt{explainability} & Technical writer & Text, score, explanation & Four-part reasoning report \\
\texttt{human\_in\_the\_loop} & Case officer & Text, last score & Case summary for a reviewer \\
\bottomrule
\end{tabular}
\caption{The roles played by the single language model in \agent{}. Each step reads its own pair of system and user prompt files.}
\label{tab:llm-roles}
\end{table}

\subsubsection{Retrieval Pipeline}
\label{sec:rag}

Indexing is done offline, once per domain. Each source PDF is parsed page by page and split with a recursive character splitter into chunks of 800 characters with an overlap of 150 characters. Each chunk is embedded with the \texttt{all-mpnet-base-v2} sentence encoder and stored in a persistent Chroma collection. The thirteen instruments of the corpus yield 847 chunks. Similarity between a query embedding $\mathbf{q}$ and a chunk embedding $\mathbf{d}$ is the cosine
\begin{equation}
\mathrm{sim}(\mathbf{q}, \mathbf{d}) = \frac{\mathbf{q}^{\top}\mathbf{d}}{\lVert \mathbf{q} \rVert \, \lVert \mathbf{d} \rVert}.
\label{eq:cosine}
\end{equation}

At analysis time, the text under evaluation is not used as a retrieval query. The context injected into the scoring prompt is the distilled summary $m$ described in Section~\ref{sec:memory}, which is the same for every text. The only place where a text is embedded and matched against the index is \texttt{compliance\_label}, which retrieves its three nearest chunks and passes the first 300 characters of each to the labeling prompt. The retrieval quality measured in Section~\ref{sec:results} is therefore the quality of the index queried directly, and not that of a per-text retrieval inside the scoring step.

\subsubsection{Memory Architecture}
\label{sec:memory}

We describe the state of the agent with the taxonomy of \citet{sumers2024cognitive}. Table~\ref{tab:memory} summarizes it, and the paragraphs below give the details.

\begin{table}[H]
\centering
\small
\begin{tabular}{@{}p{2.7cm}p{3.1cm}p{5.5cm}p{2.7cm}@{}}
\toprule
Memory type & Component & Content & Lifetime \\
\midrule
Semantic & Vector index & 847 embedded chunks of the thirteen legal instruments & Persisted on disk, built once \\
Semantic, distilled & \texttt{long\_term\_memory} & LLM summary $m$ of the 3 passages nearest to a fixed query & Computed at construction, then fixed \\
Working & \texttt{last\_score}, current text, attempt counter & Latest score and the text being rewritten & One call, but the score attribute is overwritten by every \texttt{analyze} \\
Procedural & Prompt files & Twelve templates per domain: scoring, corpus summary, healing, label, explanation, escalation & Static files on disk \\
Episodic & None & Nothing is recorded across cases, and the escalation record is returned and printed, not stored & Not applicable \\
\bottomrule
\end{tabular}
\caption{Memory of \agent{}, organized with the vocabulary of \citet{sumers2024cognitive}.}
\label{tab:memory}
\end{table}

\paragraph{Semantic memory.} The persisted vector index is the store of knowledge of the agent about the reference corpus. On top of it, the constructor computes a distilled form. It retrieves the three chunks nearest to a fixed query, whose wording is ``what are the key points of this document?'', and asks the model, with a dedicated system prompt, to extract the most useful information from them. Formally, if $q_0$ is that fixed query and $d^{(1)},d^{(2)},d^{(3)}$ are the chunks with the highest similarity to it under Equation~\ref{eq:cosine}, then
\begin{equation}
m = \mathrm{LLM}\bigl(S_{\mathrm{rag}},\; d^{(1)} \,\Vert\, d^{(2)} \,\Vert\, d^{(3)}\bigr),
\label{eq:memory}
\end{equation}
where $S_{\mathrm{rag}}$ is the summarization system prompt and the double bar denotes concatenation. The summary $m$ is stored in \texttt{self.long\_term\_memory} and never updated. Every later scoring call receives the same $m$. This design is cheap, since it costs one model call per lifetime of the agent, but it also means that the context seen by the model is generic. It is not specific to the provision that a given text concerns.

\paragraph{Working memory.} State that lives for the duration of a call consists of the most recent score, the text currently being rewritten, and the attempt counter of the self-healing loop. The loop rescores the current version of the text at each iteration and not the original one, so that the stopping test reflects the state of the correction. The score attribute is shared across calls, which means that an agent object should not be shared between concurrent requests.

\paragraph{Procedural memory.} The behavior of the agent is largely fixed by its prompt files. Each domain has its own directory of twelve templates: scoring (system and user), corpus summarization (system and user), and one system prompt and one user prompt for each of the four corrective steps. Changing the corpus therefore means changing documents and prompts, not code.

\paragraph{No episodic memory.} The agent does not remember previous cases. Each text is judged on its own, and the escalation record, which contains the score, a case summary, a priority, and the status \texttt{pending\_human\_review}, is returned to the caller and printed. Persisting it to a queue is left as future work in the code.

\subsubsection{The Corrective Action Chain}
\label{sec:action-chain}

\begin{figure}[H]
\centering
\begin{tikzpicture}[
  node distance=0.5cm and 1.1cm,
  box/.style={draw, rounded corners, align=center, minimum height=0.8cm, minimum width=3cm, font=\small},
  decision/.style={draw, diamond, aspect=2.4, align=center, font=\small, inner sep=1pt},
  arrow/.style={-{Latex[length=2mm]}}
]
\node[box] (input) {Input text $x$};
\node[box, below=of input] (analyze) {\texttt{analyze}: prompt with summary $m$\\ $\rightarrow$ score $s$, explanation};
\node[decision, below=of analyze] (decision) {$s \le \tau$ ?};
\node[box, left=1.3cm of decision] (end) {No action};
\node[box, below=0.7cm of decision] (heal) {\texttt{self\_healing}: score current text,\\ rewrite, at most 3 times};
\node[box, below=of heal] (rescore) {Rescore the final text};
\node[box, below=of rescore] (label) {\texttt{compliance\_label}\\ then \texttt{explainability}};
\node[decision, below=of label] (esc) {escalated ?};
\node[box, right=1.3cm of esc] (hitl) {\texttt{human\_in\_the\_loop}\\ priority from Eq.~\ref{eq:priority}};
\node[box, below=0.7cm of esc] (done) {Return actions taken};

\draw[arrow] (input) -- (analyze);
\draw[arrow] (analyze) -- (decision);
\draw[arrow] (decision) -- node[above, font=\small] {no} (end);
\draw[arrow] (decision) -- node[right, font=\small] {yes} (heal);
\draw[arrow] (heal) -- (rescore);
\draw[arrow] (rescore) -- (label);
\draw[arrow] (label) -- (esc);
\draw[arrow] (esc) -- node[above, font=\small] {yes} (hitl);
\draw[arrow] (esc) -- node[right, font=\small] {no} (done);
\draw[arrow] (hitl) |- (done);
\end{tikzpicture}
\caption{Control flow of \texttt{act}. A passing score ends the pipeline. A failing score enters the self-healing loop, after which the final text is rescored, labeled, and explained, and escalated to a human reviewer if the loop ended without a passing score.}
\label{fig:pipeline}
\end{figure}
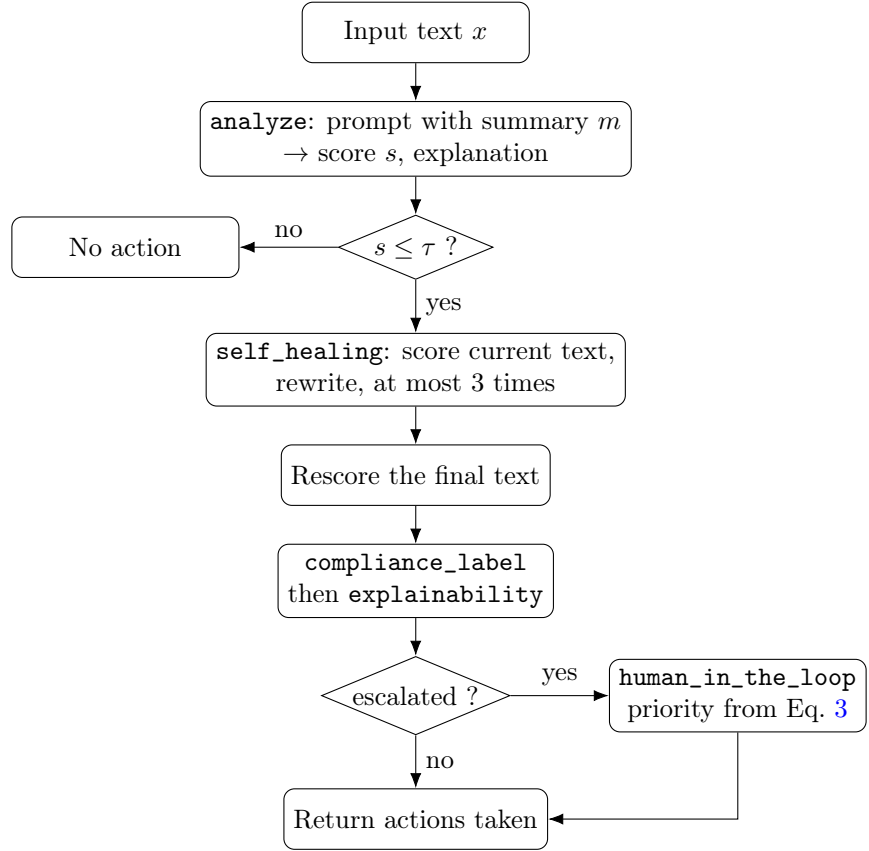

When a text scores at or below $\tau$, \texttt{act} runs four steps, shown in Figure~\ref{fig:pipeline}.

\paragraph{Self-healing.} The loop runs for at most $r=3$ iterations. Each iteration scores the current text, and if the score exceeds $\tau$ the loop stops. Otherwise the model receives the current text together with the explanation of the failure and returns a rewrite, which becomes the current text. If three rewrites were produced and the latest recorded score is still at or below $\tau$, the case is marked as escalated.

\paragraph{Compliance label.} The final text, rescored once more after the loop, is passed to a labeling step that retrieves three source excerpts and asks the model for a short structured summary: the dimensions evaluated, the score, a confidence level, and the sources.

\paragraph{Explainability.} A separate step asks for a four-part justification aimed at a non-specialist reader, covering what was evaluated, the relevant legal points, the reasoning behind the score, and the conclusion with its practical consequences.

\paragraph{Human escalation.} If the case was marked as escalated, the model drafts a case summary, and a priority is assigned by a simple rule on the last recorded score $s_{\mathrm{last}}$:
\begin{equation}
\mathrm{priority} =
\begin{cases}
\text{high} & \text{if } s_{\mathrm{last}} < \tau/2,\\
\text{medium} & \text{otherwise.}
\end{cases}
\label{eq:priority}
\end{equation}
With $\tau = 0.80$ the boundary is $0.40$.

\paragraph{Cost.} Because each step is a model call, the cost of \texttt{act} is bounded. A text that passes costs a single call. A flagged text costs between 5 and 11 calls: one scoring call in \texttt{act}, one to six calls inside the loop, made of a scoring call per iteration plus one rewrite per failed iteration, then a final scoring call, the label, the explanation, and, in the worst case, the escalation note. The worst case is $1 + 6 + 4 = 11$. The first scoring call inside the loop repeats the one made by \texttt{act} on the same text, which is a small redundancy that could be removed.

\subsubsection{Domain Configurability}
\label{sec:domain}

The constructor takes a \texttt{domain} argument that selects the reference corpus. For a given domain, the agent looks for its prompts, its persisted index, and its documents in a subdirectory of a single compliance directory, for example \texttt{compliance/legal/}. This keeps every corpus under one owned location of the project. If the domain also contains a \texttt{data/} directory of individual PDFs, each file is chunked separately and all chunks are merged into one index. The corpus of this study uses this mechanism to index thirteen instruments one by one, so that each one stays identifiable as a source. A domain without such a directory falls back to a single reference PDF. Other corpora can be added in the same way, without modifying the logic of the agent.

\subsection{Language Model and Prompting}
\label{sec:llm-prompting}

\subsubsection{Language Model Backend}
\label{sec:llm-backend}

All model calls go through a thin adapter to a hosted inference service \citep{groq}, which serves the open-weight model \texttt{openai/gpt-oss-120b} \citep{openai2025gptoss}. The adapter issues plain HTTP chat-completion requests, with temperature $0.2$ and a cap of 512 generated tokens, and exposes the same \texttt{generate} interface as the local model adapter of the framework, so the agent code does not depend on the provider. The model identifier is read from configuration. This turned out to matter in practice: the model used at the start of the development was withdrawn by its provider, and the agent was redirected to the current one by changing one configuration value, without touching any prompt.

A hosted service enforces rate limits, and a flagged text can trigger up to 11 calls in a row. When the service answers with an HTTP 429 status, the adapter waits and retries, up to five times. When the service provides a \texttt{Retry-After} header, the adapter waits for the duration it gives. Otherwise it applies an exponential backoff capped at 30 seconds, so that the waiting time before the $i$-th retry is
\begin{equation}
w_i = \min\bigl(2^{\,i-1},\, 30\bigr) \ \text{seconds}, \qquad i = 1,\dots,5.
\label{eq:backoff}
\end{equation}
\noindent The five waiting times are therefore 1, 2, 4, 8, and 16 seconds.

\subsubsection{Prompting Scheme}
\label{sec:prompting}

The scoring prompt consists of a fixed system prompt $S$ and a user template $U$ with two placeholders. Writing $x$ for the text and $m$ for the corpus summary, the user prompt is
\begin{equation}
U(x) = U\bigl[\{\mathrm{RAG1}\} \mapsto m,\; \{\mathrm{keyword1}\} \mapsto x\bigr],
\qquad
s = \pi\bigl(\mathrm{LLM}(S, U(x))\bigr),
\label{eq:prompt}
\end{equation}
where $\pi$ is the JSON extraction described in Section~\ref{sec:llm-role}. With retrieval switched off, $m$ is replaced by the string \texttt{*No RAG*}.

The system prompt of the legal domain presents the model as an expert in international legal compliance covering international law, fundamental rights, data protection, international trade, and sanctions. It anchors the scale on five reference values defined by the violations found in the text: $1.0$ for a perfectly conforming text with no identifiable violation and principles explicitly respected, $0.75$ for a mostly conforming text with minor gaps or imprecisions, $0.5$ for partial compliance where positive elements coexist with clear breaches, $0.25$ for a mostly non-conforming text with a significant violation and some mitigating elements, and $0$ for a grave violation such as an explicit rejection of international law or fundamental rights, or the use of disproportionate coercive measures. It also asks for intermediate values, with examples such as $0.95$, $0.53$, and $0.20$, and forbids assuming any earlier evaluation. The prompts are written in French, the model is told to answer in French, and the texts to evaluate are in French. This choice follows from the local setting of the project: in Quebec, the documents to be checked are written in French, and the thirteen reference documents are French versions as well. The rest of the framework, including the ethics sub-agent, is written in English from end to end, so the use of French in this sub-agent is a deliberate adaptation to the Quebec context and not a property of the framework.

\subsection{Data}
\label{sec:data}

\subsubsection{Reference Corpus}

The reference corpus brings together thirteen legal instruments chosen to reflect the Canadian and Quebec context in which the project is carried out. It combines federal texts, Quebec statutes, documents of the City of Trois-Rivi\`eres, and international instruments that frame them (Table~\ref{tab:corpus}). The municipal documents are the employee code of ethics, the privacy policy, and the normative urban planning by-law of the City of Trois-Rivi\`eres. Each instrument is indexed as a separate PDF, and the thirteen files together yield 847 chunks.

\begin{table}[H]
\centering
\small
\begin{tabular}{@{}p{3.2cm}p{9.8cm}@{}}
\toprule
Level & Instruments \\
\midrule
Canada (3) & Chapter 19 of the Canada-United States-Mexico Agreement on digital trade \citep{cusma2020ch19}; Canadian Charter of Rights and Freedoms \citep{canada1982charter}; Personal Information Protection and Electronic Documents Act \citep{canada2000pipeda} \\
Quebec (4) & Law~25 modernizing the protection of personal information \citep{quebec2021loi25}; Charter of Human Rights and Freedoms \citep{quebec1975charter}; Act respecting access to documents held by public bodies \citep{quebec1982access}; Environment Quality Act \citep{quebec1972environment} \\
City of Trois-Rivi\`eres (3) & Employee code of ethics; privacy policy; normative urban planning by-law \citep{troisrivieres} \\
International (3) & Charter of the United Nations \citep{un1945charter}; International Covenant on Civil and Political Rights \citep{un1966iccpr}; Paris Agreement \citep{unfccc2015paris} \\
\bottomrule
\end{tabular}
\caption{The thirteen instruments of the reference corpus, grouped by level of government.}
\label{tab:corpus}
\end{table}

\subsubsection{Gold Standard}

The gold standard contains fifteen French texts, organized in three annotated categories of five texts each: conforming texts (C1 to C5), texts with exactly one violation (V1 to V5), and texts with several violations (M1 to M5). Every text carries a conformity label, the list of expected violation types, and an expected score interval: $[0.65, 1.0]$ for conforming texts, $[0, 0.45]$ for texts with one violation, and an upper bound between $0.25$ and $0.35$ for texts with several violations. The violation types are fundamental rights, international trade, data protection, sanctions, and international law. Table~\ref{tab:examples} gives English translations of one text per category.

The texts are fictional diplomatic and governmental statements involving invented states. They were generated with a large language model, and we then annotated them manually. They were not tuned to the present agent.

\begin{table}[H]
\centering
\small
\begin{tabular}{@{}p{0.7cm}p{3.0cm}p{9.3cm}@{}}
\toprule
ID & Category & Text (English translation of the French original) \\
\midrule
C3 & Conforming & Under the Paris Agreement, our nation commits to cutting its CO$_2$ emissions by 45\% by 2030. We propose an independent and transparent verification mechanism open to all signatory states. International climate cooperation is an absolute priority of our foreign policy. \\
V4 & One violation, sanctions & We unilaterally impose a total embargo on all medicine exports to Vostaria, including essential medicines intended for civilian populations. The measure is immediate and without appeal. \\
M4 & Several violations, international law, fundamental rights, data protection & If Solane does not sign the trade agreement within 48 hours, we will launch cyberattacks on its infrastructure. We will also disclose the personal data of all Solanese diplomats. Moreover, we will block all humanitarian access to populations in crisis at the border. \\
\bottomrule
\end{tabular}
\caption{One example text for each of the three annotated categories of the gold standard.}
\label{tab:examples}
\end{table}

\subsubsection{Retrieval Test Queries}

Five queries were written with lists of relevant keywords. They cover digital trade, the Paris Agreement, the United Nations Charter, the protection of the family and fundamental rights, and the protection of personal information. They are written in French, except the first, which is in English.

\subsection{Evaluation Protocol and Metrics}
\label{sec:evalmetrics}

\subsubsection{Protocol}
\label{sec:protocol}

The evaluation is organized in five blocks, implemented in a notebook. The first block loads the agent and the gold standard. The second runs the stability test, in which the three texts C1, V1, and M1 are each scored five times. The third scores each of the fifteen texts once and computes the metrics of Section~\ref{sec:metrics} at $\tau = 0.80$. The fourth runs the retrieval queries against the index. The fifth computes the composite score and exports a JSON report. Calls are spaced by one second, and the temperature is $0.2$ throughout. The results below come from the archived report dated 16 September 2026. Only the scoring step is evaluated here. The corrective chain of \texttt{act} is described but not scored, since evaluating rewrites would require a second gold standard.

\subsubsection{Decision Rule and Classification Metrics}
\label{sec:metrics}

Let the gold standard contain $N$ texts, each with a label $y_i \in \{0,1\}$ where $y_i = 1$ means non-conforming, and let $s_i$ be the score returned for text $i$. The predicted label is
\begin{equation}
\hat{y}_i =
\begin{cases}
1 & \text{if } s_i \le \tau \text{ or no score was returned},\\
0 & \text{otherwise,}
\end{cases}
\label{eq:decision}
\end{equation}
so that a failed analysis counts as flagged, which is the conservative choice for a compliance tool. From the counts of true positives $\mathrm{TP}$, false positives $\mathrm{FP}$, true negatives $\mathrm{TN}$, and false negatives $\mathrm{FN}$, we compute
\begin{equation}
P = \frac{\mathrm{TP}}{\mathrm{TP}+\mathrm{FP}}, \quad
R = \frac{\mathrm{TP}}{\mathrm{TP}+\mathrm{FN}}, \quad
F_1 = \frac{2PR}{P+R}, \quad
\mathrm{Acc} = \frac{\mathrm{TP}+\mathrm{TN}}{N}, \quad
\mathrm{Spec} = \frac{\mathrm{TN}}{\mathrm{TN}+\mathrm{FP}}.
\label{eq:prf}
\end{equation}
Each text also carries an expected score interval $[\ell_i, u_i]$ chosen by the annotator, and the calibration rate is the fraction of scores that fall inside it,
\begin{equation}
\kappa = \frac{\bigl|\{\, i : \ell_i \le s_i \le u_i \,\}\bigr|}{N}.
\label{eq:calibration}
\end{equation}
With so few texts, point estimates are fragile. We therefore attach Wilson score intervals \citep{wilson1927probable} to proportions. For $k$ successes out of $n$, with $\hat p = k/n$ and $z=1.96$, the interval is
\begin{equation}
\frac{\hat p + \dfrac{z^2}{2n} \pm z\sqrt{\dfrac{\hat p(1-\hat p)}{n} + \dfrac{z^2}{4n^2}}}{1+\dfrac{z^2}{n}}.
\label{eq:wilson}
\end{equation}

\subsubsection{Stability}

Because the model is sampled at a nonzero temperature, the same text can receive different scores. For a representative text $j$ we submit it $R=5$ times and compute the sample standard deviation
\begin{equation}
\sigma_j = \sqrt{\frac{1}{R-1}\sum_{r=1}^{R}\bigl(s_{jr}-\bar{s}_j\bigr)^2},
\qquad
\bar\sigma = \frac{1}{J}\sum_{j=1}^{J}\sigma_j,
\label{eq:sigma}
\end{equation}
over the $J=3$ test texts. The agent is deemed stable if $\bar\sigma < 0.10$, a limit that we fixed in advance.

\subsubsection{Retrieval Quality}

For a query $q$ we take the three chunks nearest to it in the index and mark each as relevant if its lowercased text contains at least one keyword from a list written for that query. With $\mathrm{rel}_k(q)$ equal to $1$ if the $k$-th passage is relevant and $0$ otherwise, precision at three and its mean over $Q$ queries are
\begin{equation}
\mathrm{P@3}(q) = \frac{1}{3}\sum_{k=1}^{3}\mathrm{rel}_k(q),
\qquad
\overline{\mathrm{P@3}} = \frac{1}{Q}\sum_{q}\mathrm{P@3}(q).
\label{eq:p3}
\end{equation}

\subsubsection{Composite Score}

We combine the four blocks into a single number with weights that we fixed in advance and did not fit to the results,
\begin{equation}
S = 0.40\,F_1 + 0.25\,\bigl(1-5\bar\sigma\bigr) + 0.20\,\overline{\mathrm{P@3}} + 0.15\,\kappa,
\label{eq:composite}
\end{equation}
clipped to $[0,1]$. The factor $5$ maps the stability limit $\bar\sigma = 0.10$ to half of the range of the stability term. The composite is a convenience for comparing runs. It is not a validated measure, and we report its components alongside it.

\subsubsection{Supplementary Analyses}

Two analyses were added after seeing the first results, and we label them as post hoc. The first is the area under the ROC curve \citep{fawcett2006roc}, which does not depend on $\tau$ and measures only whether non-conforming texts receive lower scores than conforming ones. With $\mathcal{P}$ the set of non-conforming texts and $\mathcal{N}$ the set of conforming ones, it is the average over all pairs $(p,n)$ of the pair score $c(p,n)$,
\begin{equation}
\mathrm{AUC} = \frac{1}{|\mathcal{P}|\,|\mathcal{N}|}\sum_{p\in\mathcal{P}}\sum_{n\in\mathcal{N}} c(p,n),
\qquad
c(p,n) =
\begin{cases}
1 & \text{if } s_n > s_p,\\
1/2 & \text{if } s_n = s_p,\\
0 & \text{if } s_n < s_p.
\end{cases}
\label{eq:auc}
\end{equation}
The second is a sweep of the threshold $\tau$ from $0.05$ to $0.95$ in steps of $0.05$, recomputing Equation~\ref{eq:prf} at each value, to see how sensitive the decision is to the one parameter that was set in advance.

\section{Results}
\label{sec:results}

\subsection{Classification}

Table~\ref{tab:main} gives the main results. All fifteen texts are classified correctly: the ten non-conforming texts are flagged, and the five conforming texts are not. The Wilson interval for the accuracy of $15/15$ is $[0.80, 1.00]$, which is a reminder that a perfect result on fifteen texts is compatible with a true accuracy as low as $0.80$. The same reasoning applies to the recall, with an interval of $[0.72, 1.00]$, and to the specificity, with $[0.57, 1.00]$.

\begin{table}[H]
\centering
\small
\begin{tabular}{@{}lc@{}}
\toprule
Metric & Value \\
\midrule
TP / FP / TN / FN & 10 / 0 / 5 / 0 \\
Precision & 1.000 \\
Recall (Wilson 95\% interval) & 1.000 $[0.72, 1.00]$ \\
F$_1$ & 1.000 \\
Accuracy (Wilson 95\% interval) & 1.000 $[0.80, 1.00]$ \\
Specificity (Wilson 95\% interval) & 1.000 $[0.57, 1.00]$ \\
Calibration rate $\kappa$ & 1.000 \\
AUC (post hoc) & 1.000 \\
Mean stability $\bar\sigma$ & 0.0303 \\
Mean retrieval P@3 & 1.000 \\
Composite score $S$ & 0.962 \\
\bottomrule
\end{tabular}
\caption{Main results at $\tau = 0.80$ on the gold standard of fifteen texts, of which ten are non-conforming and five are conforming.}
\label{tab:main}
\end{table}

\subsection{Scores per Text}

\begin{figure}[H]
\centering
\includegraphics[width=0.8\linewidth]{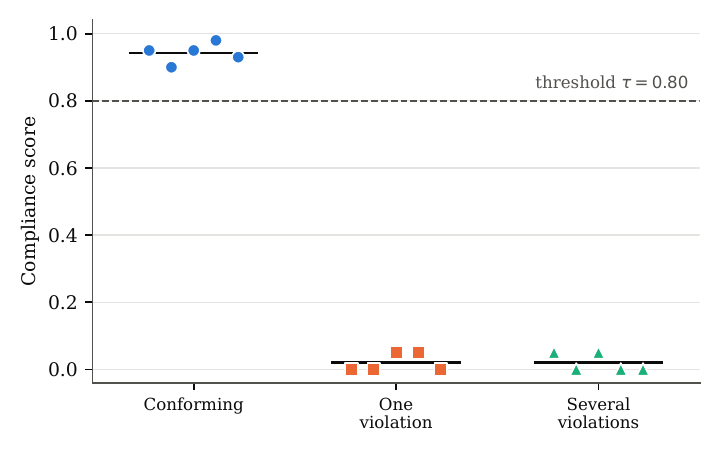}
\caption{Scores of the fifteen gold standard texts by category. Horizontal bars are category means and the dashed line is the threshold $\tau = 0.80$.}
\label{fig:scores}
\end{figure}

Table~\ref{tab:pertext} lists the score of every text, and Figure~\ref{fig:scores} shows the same information by category. The five conforming texts receive between $0.90$ and $0.98$, with a mean of $0.942$. The ten non-conforming texts receive between $0$ and $0.05$, with a mean of $0.020$ for the texts with one violation and $0.020$ for those with several. Every score lies inside the expected interval annotated for its text, which gives $\kappa = 1$.

\begin{table}[H]
\centering
\small
\begin{tabular}{@{}llp{4.8cm}cc@{}}
\toprule
ID & Category & Expected violation types & Expected bound & Score \\
\midrule
C1 & Conforming & none & at least 0.65 & 0.95 \\
C2 & Conforming & none & at least 0.65 & 0.90 \\
C3 & Conforming & none & at least 0.65 & 0.95 \\
C4 & Conforming & none & at least 0.65 & 0.98 \\
C5 & Conforming & none & at least 0.65 & 0.93 \\
V1 & One violation & fundamental rights & at most 0.45 & 0.00 \\
V2 & One violation & international trade & at most 0.45 & 0.00 \\
V3 & One violation & data protection & at most 0.45 & 0.05 \\
V4 & One violation & sanctions & at most 0.45 & 0.05 \\
V5 & One violation & international law & at most 0.45 & 0.00 \\
M1 & Several violations & international law, fundamental rights & at most 0.35 & 0.05 \\
M2 & Several violations & fundamental rights, data protection, sanctions & at most 0.30 & 0.00 \\
M3 & Several violations & international law, fundamental rights, international trade & at most 0.30 & 0.05 \\
M4 & Several violations & international law, fundamental rights, data protection & at most 0.25 & 0.00 \\
M5 & Several violations & international law, fundamental rights, international trade, sanctions & at most 0.25 & 0.00 \\
\bottomrule
\end{tabular}
\caption{Expected annotation and score of every text of the gold standard.}
\label{tab:pertext}
\end{table}

The scores are strongly polarized. The highest score of a non-conforming text is $0.05$ and the lowest score of a conforming text is $0.90$, which leaves a gap of $0.85$. The post hoc AUC is therefore $1$, and the threshold sweep gives the same perfect classification for every tested value from $\tau = 0.05$ to $\tau = 0.85$. At $\tau = 0.90$ the text C2, which scores exactly $0.90$, is flagged, and precision falls to $0.91$. The choice of $\tau = 0.80$ is thus not delicate on this gold standard, but this also means that the gold standard cannot tell us how the threshold would behave on texts whose scores fall in the middle of the scale. This is exactly the question raised by borderline cases: how does the method react when a text is only partly conforming or ambiguous? The present data cannot answer it, and we list it among the first priorities in Section~\ref{sec:discussion}.

\subsection{Stability}

\begin{figure}[H]
\centering
\includegraphics[width=0.8\linewidth]{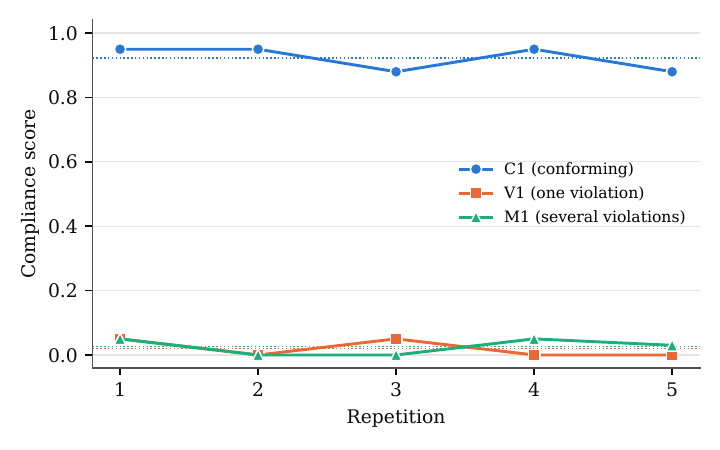}
\caption{Scores over five repetitions of the same text. Dotted lines are the means of each series.}
\label{fig:stab}
\end{figure}

Table~\ref{tab:stab} reports the five repetitions of the three test texts. The mean standard deviation is $\bar\sigma = 0.0303$, well below the limit of $0.10$. Variation is the largest on the conforming text, which ranges from $0.88$ to $0.95$, and the two other texts stay between $0$ and $0.05$. Figure~\ref{fig:stab} shows the repetitions.

\begin{table}[H]
\centering
\small
\begin{tabular}{@{}llccc@{}}
\toprule
Text & Category & Five scores & Mean & $\sigma_j$ \\
\midrule
C1 & Conforming & 0.95, 0.95, 0.88, 0.95, 0.88 & 0.922 & 0.038 \\
V1 & One violation & 0.05, 0.00, 0.05, 0.00, 0.00 & 0.020 & 0.027 \\
M1 & Several violations & 0.05, 0.00, 0.00, 0.05, 0.03 & 0.026 & 0.025 \\
\bottomrule
\end{tabular}
\caption{Stability test, with five repetitions of the same text. The mean over the three texts is $\bar\sigma = 0.0303$.}
\label{tab:stab}
\end{table}

\subsection{Retrieval}

Mean precision at three is $1.000$ (Table~\ref{tab:rag}): for each of the five queries, the three nearest chunks all contain at least one of the keywords written for that query. This result needs a careful reading. The keywords are generic words such as ``droit'' (law, right) or ``protection'' (protection), which occur in most passages of a legal corpus, so a score of $1.00$ is a weak signal of relevance. It shows that the index returns legal passages that are on topic at a coarse level, and not that it returns the right article.

\begin{table}[H]
\centering
\small
\begin{tabular}{@{}p{10.5cm}c@{}}
\toprule
Query (abbreviated and translated from French when needed) & P@3 \\
\midrule
digital trade algorithm computing facility & 1.00 \\
Paris Agreement, emissions reduction, climate commitment & 1.00 \\
UN Charter, international law, sovereignty & 1.00 \\
protection of the family and fundamental rights of the person & 1.00 \\
protection of personal information, consent, confidentiality & 1.00 \\
\midrule
Mean & 1.000 \\
\bottomrule
\end{tabular}
\caption{Precision at three of the vector index for the five test queries.}
\label{tab:rag}
\end{table}

\section{Discussion and Perspectives}
\label{sec:discussion}

\subsection{What the Results Say}

On this gold standard the agent separates the two classes without error and with a wide margin. Conforming texts score above $0.90$ and the others at or below $0.05$. The agent is also stable, since five repetitions of the same text spread by at most $0.07$ for the conforming text and by $0.05$ for the others. Together with a retrieval step that returns on-topic passages, this shows that the pipeline works as designed on the cases it was tested on: the model produces parsable scores, the threshold rule flags the right texts, and nothing in the control flow breaks.

\paragraph{Answers to the research questions.} On design, the architecture separates the roles of the model, of the index, and of the working state, and the single JSON contract on the score keeps the pipeline testable. The generic corpus summary shows, however, that the contribution of retrieval at the moment of judgment is limited. On performance, the agent separated every conforming text from every non-conforming one on this gold standard, with stable scores. On evidence, fifteen texts, one annotator, and blunt violations support the claim that the pipeline works, but not a measure of accuracy in realistic conditions.

\subsection{How Far These Results Go}

The perfect scores should not be read as a measure of accuracy in realistic conditions, and we want to say why. First, the non-conforming statements are blunt. Most of them declare outright that international law or fundamental rights do not apply, or announce coercive measures, so that a language model may recognize them without consulting the corpus at all. Second, the conforming statements are uniformly positive, and several of them name instruments that belong to the corpus, such as the United Nations Charter, the Paris Agreement, Law~25, and PIPEDA, which is a strong cue. Third, the states are fictional and the texts are short. Fourth, the scale is used as a binary one. The intermediate anchors $0.25$, $0.5$, and $0.75$ are never used: every observed score is either at most $0.05$ or at least $0.88$. We therefore have no evidence on how the agent grades a partially conforming text, or a statement that is ambiguous, which are the cases where a compliance tool is the most needed. Finally, the evaluation includes no baseline without retrieval, although the agent supports one, so we cannot attribute any part of the result to retrieval.

\subsection{Limits of the Evaluation}

The conclusions are bounded by several limitations. The gold standard contains fifteen texts, so the intervals are wide, and a perfect score is compatible with an accuracy near $0.80$. We assigned the labels and the expected score intervals ourselves, with no independent annotator, so we have no measure of inter-annotator agreement, and the calibration rate is partly circular. The stability test covers three texts only, and one model, hosted by a third party, was used throughout, whose version may change. The relevance judgments for retrieval are keyword based and lenient, as discussed. The evaluation covers a single corpus and a single formulation of the rubric, and we make no claim that the results carry over to other regulatory corpora.

Two points concern the implementation. The escalation flag of the self-healing loop is computed from the last score recorded before the final rewrite, so a final rewrite that would have passed is still marked as escalated, and the first scoring call of the loop duplicates the one made by \texttt{act}. Neither affects the metrics of this paper, since only the scoring step was evaluated, but both should be corrected before the corrective chain is itself evaluated.

\subsection{Perspectives}

The validation presented here is enough to show that the pipeline works, but a journal version needs a stronger one. The test data should be more numerous, documented, and include borderline cases, and annotation by several people would allow agreement to be measured. Stability runs on more texts and an ablation without retrieval would complete it. Further ablations should compare the current design with a per-text retrieval and with a summary built from several queries, and real documents such as municipal and provincial policies could be added to the texts. An embedding model suited to French, and a chunking that follows the structure of articles, could improve the retrieval step. The corrective chain needs its own evaluation, including whether a rewrite preserves the meaning of the original text. Finally, an episodic memory that stores the decisions of reviewers on escalated cases would let the agent adjust to the practice of the people who use it. A separate study should then apply the method in a real setting, with the documents of the City of Trois-Rivi\`eres and the work carried out with it, as an applied study of the compliance sub-agent.

\section{Conclusion}
\label{sec:conclusion}

We presented \agent{}, a compliance sub-agent for \compass{} in which one language model scores, rewrites, labels, explains, and escalates, supported by a persisted vector index and a one-time corpus summary, and configurable by domain. We described what the retrieval step does and does not do at analysis time, we mapped the memory of the agent to a standard taxonomy while noting what it lacks, and we defined the evaluation metrics. On a corpus of thirteen Canadian, Quebec, municipal, and international instruments, the agent classified fifteen of fifteen annotated texts correctly with stable scores. The size and the bluntness of the gold standard limit what this result proves, and the next step is a larger and better documented evaluation, with borderline texts and annotation by several people, followed by an applied study in a real setting with the City of Trois-Rivi\`eres. The evaluation notebook, the prompts, and the JSON report are part of the project and are available from us on request.

\section*{Acknowledgments}

We thank the members of the \compass{} research team for their discussions about the framework. The sovereignty, carbon, and ethics sub-agents are developed by other team members and are not part of this work.

\section*{Use of AI Tools}

We used AI tools to check and improve the language of this manuscript. The research design, the experiments, the results, and their interpretation are our own work, and we take full responsibility for the content.

\bibliographystyle{plainnat}
\bibliography{references}

\begin{thebibliography}{32}
\providecommand{\natexlab}[1]{#1}
\providecommand{\url}[1]{\texttt{#1}}
\expandafter\ifx\csname urlstyle\endcsname\relax
  \providecommand{\doi}[1]{doi: #1}\else
  \providecommand{\doi}{doi: \begingroup \urlstyle{rm}\Url}\fi

\bibitem[Chalkidis et~al.(2020)Chalkidis, Fergadiotis, Malakasiotis, Aletras,
  and Androutsopoulos]{chalkidis2020legalbert}
Ilias Chalkidis, Manos Fergadiotis, Prodromos Malakasiotis, Nikolaos Aletras,
  and Ion Androutsopoulos.
\newblock {LEGAL-BERT}: The muppets straight out of law school.
\newblock In \emph{Findings of the Association for Computational Linguistics:
  EMNLP 2020}. Association for Computational Linguistics, 2020.

\bibitem[{Chroma}(2023)]{chromadb}
{Chroma}.
\newblock Chroma: The open-source embedding database.
\newblock \url{https://www.trychroma.com}, 2023.

\bibitem[Dahl et~al.(2024)Dahl, Magesh, Suzgun, and Ho]{dahl2024legalfictions}
Matthew Dahl, Varun Magesh, Mirac Suzgun, and Daniel~E. Ho.
\newblock Large legal fictions: Profiling legal hallucinations in large
  language models.
\newblock \emph{Journal of Legal Analysis}, 16\penalty0 (1), 2024.

\bibitem[Dessureault et~al.(2026)Dessureault, Iliho~Manzi, Alaoui~Ismaili, Lo,
  Lalancette, and B{\'e}langer]{dessureault2026compass}
Jean-S{\'e}bastien Dessureault, Alain-Thierry Iliho~Manzi, Soukaina
  Alaoui~Ismaili, Khadim Lo, Mireille Lalancette, and {\'E}ric B{\'e}langer.
\newblock {COMPASS}: The explainable agentic framework for sovereignty,
  sustainability, compliance, and ethics.
\newblock arXiv preprint arXiv:2603.11277, 2026.

\bibitem[Fawcett(2006)]{fawcett2006roc}
Tom Fawcett.
\newblock An introduction to {ROC} analysis.
\newblock \emph{Pattern Recognition Letters}, 27\penalty0 (8), 2006.

\bibitem[Gao et~al.(2023)Gao, Xiong, Gao, Jia, Pan, Bi, Dai, Sun, Wang, and
  Wang]{gao2023ragsurvey}
Yunfan Gao, Yun Xiong, Xinyu Gao, Kangxiang Jia, Jinliu Pan, Yuxi Bi, Yi~Dai,
  Jiawei Sun, Meng Wang, and Haofen Wang.
\newblock Retrieval-augmented generation for large language models: A survey,
  2023.

\bibitem[{Gouvernement du Qu{\'e}bec}(1972)]{quebec1972environment}
{Gouvernement du Qu{\'e}bec}.
\newblock {Environment Quality Act, CQLR c.~Q-2}, 1972.

\bibitem[{Gouvernement du Qu{\'e}bec}(1975)]{quebec1975charter}
{Gouvernement du Qu{\'e}bec}.
\newblock {Charter of Human Rights and Freedoms, CQLR c.~C-12}, 1975.

\bibitem[{Gouvernement du Qu{\'e}bec}(1982)]{quebec1982access}
{Gouvernement du Qu{\'e}bec}.
\newblock {Act Respecting Access to Documents Held by Public Bodies and the
  Protection of Personal Information, CQLR c.~A-2.1}, 1982.

\bibitem[{Gouvernement du Qu{\'e}bec}(2021)]{quebec2021loi25}
{Gouvernement du Qu{\'e}bec}.
\newblock {An Act to Modernize Legislative Provisions as Regards the Protection
  of Personal Information (Loi 25)}, 2021.

\bibitem[{Government of Canada}(1982)]{canada1982charter}
{Government of Canada}.
\newblock {Canadian Charter of Rights and Freedoms, Part I of the Constitution
  Act, 1982}, 1982.

\bibitem[{Government of Canada}(2000)]{canada2000pipeda}
{Government of Canada}.
\newblock {Personal Information Protection and Electronic Documents Act
  (PIPEDA)}, 2000.

\bibitem[{Government of Canada}(2020)]{cusma2020ch19}
{Government of Canada}.
\newblock {Canada-United States-Mexico Agreement, Chapter 19: Digital Trade},
  2020.

\bibitem[{Groq, Inc.}(2024)]{groq}
{Groq, Inc.}
\newblock Groq {LPU} inference engine.
\newblock \url{https://groq.com}, 2024.

\bibitem[Guha et~al.(2023)Guha, Nyarko, Ho, R{\'e}, et~al.]{guha2023legalbench}
Neel Guha, Julian Nyarko, Daniel~E. Ho, Christopher R{\'e}, et~al.
\newblock {LegalBench}: A collaboratively built benchmark for measuring legal
  reasoning in large language models.
\newblock In \emph{Advances in Neural Information Processing Systems, Datasets
  and Benchmarks Track}, volume~36, 2023.

\bibitem[{International Organization for Standardization}(2023)]{iso42001}
{International Organization for Standardization}.
\newblock {ISO/IEC} 42001:2023, information technology: Artificial intelligence
  management system, 2023.

\bibitem[{LangChain, Inc.}(2023)]{langchain}
{LangChain, Inc.}
\newblock {L}ang{C}hain: Building applications with {LLM}s through
  composability.
\newblock \url{https://www.langchain.com}, 2023.

\bibitem[Lewis et~al.(2020)Lewis, Perez, Piktus, Petroni, Karpukhin, Goyal,
  K{\"u}ttler, Lewis, Yih, Rockt{\"a}schel, Riedel, and
  Kiela]{lewis2020retrieval}
Patrick Lewis, Ethan Perez, Aleksandra Piktus, Fabio Petroni, Vladimir
  Karpukhin, Naman Goyal, Heinrich K{\"u}ttler, Mike Lewis, Wen-tau Yih, Tim
  Rockt{\"a}schel, Sebastian Riedel, and Douwe Kiela.
\newblock Retrieval-augmented generation for knowledge-intensive {NLP} tasks.
\newblock In \emph{Advances in Neural Information Processing Systems},
  volume~33, 2020.

\bibitem[Madaan et~al.(2023)Madaan, Tandon, Gupta, Hallinan, Gao, Wiegreffe,
  Alon, Dziri, Prabhumoye, Yang, Gupta, Majumder, Hermann, Welleck,
  Yazdanbakhsh, and Clark]{madaan2023selfrefine}
Aman Madaan, Niket Tandon, Prakhar Gupta, Skyler Hallinan, Luyu Gao, Sarah
  Wiegreffe, Uri Alon, Nouha Dziri, Shrimai Prabhumoye, Yiming Yang, Shashank
  Gupta, Bodhisattwa~Prasad Majumder, Katherine Hermann, Sean Welleck, Amir
  Yazdanbakhsh, and Peter Clark.
\newblock Self-refine: Iterative refinement with self-feedback.
\newblock In \emph{Advances in Neural Information Processing Systems},
  volume~36, 2023.

\bibitem[Mitchell et~al.(2019)Mitchell, Wu, Zaldivar, Barnes, Vasserman,
  Hutchinson, Spitzer, Raji, and Gebru]{mitchell2019modelcards}
Margaret Mitchell, Simone Wu, Andrew Zaldivar, Parker Barnes, Lucy Vasserman,
  Ben Hutchinson, Elena Spitzer, Inioluwa~Deborah Raji, and Timnit Gebru.
\newblock Model cards for model reporting.
\newblock In \emph{Proceedings of the Conference on Fairness, Accountability,
  and Transparency}. ACM, 2019.

\bibitem[{National Institute of Standards and Technology}(2023)]{nist2023airmf}
{National Institute of Standards and Technology}.
\newblock Artificial intelligence risk management framework ({AI RMF} 1.0).
\newblock Technical Report NIST AI 100-1, U.S. Department of Commerce, 2023.

\bibitem[{OpenAI}(2025)]{openai2025gptoss}
{OpenAI}.
\newblock gpt-oss-120b and gpt-oss-20b model card.
\newblock Technical model card, 2025.

\bibitem[Reimers and Gurevych(2019)]{reimers2019sentencebert}
Nils Reimers and Iryna Gurevych.
\newblock Sentence-{BERT}: Sentence embeddings using {S}iamese {BERT}-networks.
\newblock In \emph{Proceedings of the 2019 Conference on Empirical Methods in
  Natural Language Processing and the 9th International Joint Conference on
  Natural Language Processing (EMNLP-IJCNLP)}. Association for Computational
  Linguistics, 2019.

\bibitem[Shinn et~al.(2023)Shinn, Cassano, Gopinath, Narasimhan, and
  Yao]{shinn2023reflexion}
Noah Shinn, Federico Cassano, Ashwin Gopinath, Karthik Narasimhan, and Shunyu
  Yao.
\newblock Reflexion: Language agents with verbal reinforcement learning.
\newblock In \emph{Advances in Neural Information Processing Systems},
  volume~36, 2023.

\bibitem[Song et~al.(2020)Song, Tan, Qin, Lu, and Liu]{song2020mpnet}
Kaitao Song, Xu~Tan, Tao Qin, Jianfeng Lu, and Tie-Yan Liu.
\newblock {MPN}et: Masked and permuted pre-training for language understanding.
\newblock In \emph{Advances in Neural Information Processing Systems},
  volume~33, 2020.

\bibitem[Sumers et~al.(2024)Sumers, Yao, Narasimhan, and
  Griffiths]{sumers2024cognitive}
Theodore~R. Sumers, Shunyu Yao, Karthik Narasimhan, and Thomas~L. Griffiths.
\newblock Cognitive architectures for language agents.
\newblock \emph{Transactions on Machine Learning Research}, 2024.

\bibitem[{United Nations}(1945)]{un1945charter}
{United Nations}.
\newblock {Charter of the United Nations}.
\newblock United Nations, 1945.

\bibitem[{United Nations Framework Convention on Climate
  Change}(2015)]{unfccc2015paris}
{United Nations Framework Convention on Climate Change}.
\newblock {Paris Agreement under the United Nations Framework Convention on
  Climate Change}, 2015.

\bibitem[{United Nations General Assembly}(1966)]{un1966iccpr}
{United Nations General Assembly}.
\newblock {International Covenant on Civil and Political Rights}.
\newblock United Nations Treaty Series, vol. 999, p. 171, 1966.

\bibitem[{Ville de Trois-Rivi{\`e}res}(n.d.)]{troisrivieres}
{Ville de Trois-Rivi{\`e}res}.
\newblock {Employee Code of Ethics, Privacy Policy, and Normative Urban
  Planning By-law}.
\newblock Municipal documents, n.d.

\bibitem[Wilson(1927)]{wilson1927probable}
Edwin~B. Wilson.
\newblock Probable inference, the law of succession, and statistical inference.
\newblock \emph{Journal of the American Statistical Association}, 22\penalty0
  (158), 1927.

\bibitem[Zheng et~al.(2023)Zheng, Chiang, Sheng, Zhuang, Wu, Zhuang, Lin, Li,
  Li, Xing, Zhang, Gonzalez, and Stoica]{zheng2023judging}
Lianmin Zheng, Wei-Lin Chiang, Ying Sheng, Siyuan Zhuang, Zhanghao Wu, Yonghao
  Zhuang, Zi~Lin, Zhuohan Li, Dacheng Li, Eric~P. Xing, Hao Zhang, Joseph~E.
  Gonzalez, and Ion Stoica.
\newblock Judging {LLM}-as-a-judge with {MT}-bench and {C}hatbot {A}rena.
\newblock In \emph{Advances in Neural Information Processing Systems},
  volume~36, 2023.

\end{thebibliography}

\end{document}